\documentclass[sigconf, screen, table]{acmart}
\usepackage{pifont}
\usepackage{fancyhdr}

\AtBeginDocument{%
  }

\setcopyright{acmlicensed}
\copyrightyear{2025}
\acmYear{2025}
\acmDOI{XXXXXXX.XXXXXXX}
\acmConference[Conference acronym '25]{Make sure to enter the correct
  conference title from your rights confirmation email}{October 27–31,
  2025}{Dublin, Ireland}
\acmISBN{978-1-4503-XXXX-X/2018/06}

\renewcommand\footnotetextcopyrightpermission[1]{}

\acmSubmissionID{51}

\usepackage{eso-pic}
\usepackage{graphicx}
\usepackage{hyperref} 

\usepackage{pifont}
\usepackage{graphicx}
\usepackage{wrapfig}
\usepackage{graphicx}
\usepackage{booktabs}
\usepackage{svg} 
\usepackage{colortbl}  
\usepackage{cleveref}
\usepackage{multirow}
\usepackage{indentfirst}
\usepackage{threeparttable}
\usepackage{pifont}
\usepackage{makecell}
\usepackage{arydshln}
\usepackage{arydshln}
\usepackage{tabularx}

\usepackage{makecell}

\usepackage{pifont}
\usepackage{minitoc}
\definecolor{mblue}{RGB}{0, 77, 128}

\definecolor{mblue}{RGB}{0, 77, 128}
\definecolor{mred}{RGB}{192,0, 0}
\definecolor{darkgreen}{rgb}{0.0, 0.5, 0.0}
\definecolor{mycolor_blue}{HTML}{E7EFFA}

\definecolor{mycolor_gray}{HTML}{ECECEC}

\hypersetup{
    colorlinks=true,        
    linkcolor=blue,        
    filecolor=magenta,     
    urlcolor=magenta,          
}

\begin{document}

\AddToShipoutPictureBG*{%
  \put(\LenToUnit{\dimexpr 6.5in + 1.05cm}, \LenToUnit{\dimexpr 11in - 2.65cm}){%
    \includegraphics[height=5mm]{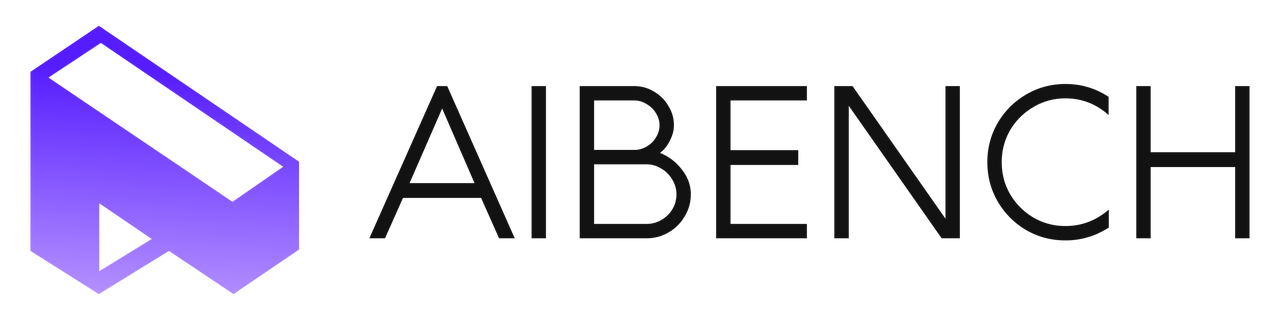}%
  }%
}

\pagestyle{fancy}
\fancyhf{}
\fancyhead[R]{\includegraphics[height=5mm]{image/AIBen.PNG}}  
\fancyfoot[C]{\thepage}  

\title{GOBench: Benchmarking Geometric Optics Generation
and Understanding of MLLMs}

\author{Xiaorong Zhu}
\email{zhuxiaorong@sjtu.edu.cn}
\affiliation{%
  \institution{Shanghai Jiao Tong University}
  \city{Shanghai}
  \country{China}
}

\author{Ziheng Jia}
\affiliation{%
  \institution{Shanghai Jiao Tong University}
  \city{Shanghai}
  \country{China}
}

\author{Jiarui Wang}
\affiliation{%
  \institution{Shanghai Jiao Tong University}
  \city{Shanghai}
  \country{China}
}

\author{Xiangyu Zhao}
\affiliation{%
  \institution{Shanghai Jiao Tong University}
  \institution{Shanghai AI Laboratory}
  \city{Shanghai}
  \country{China}
}

\author{Haodong Duan}
\affiliation{%
  \institution{Shanghai AI Laboratory}
  \city{Shanghai}
  \country{China}
}

\author{Xiongkuo Min}
\affiliation{%
  \institution{Shanghai Jiao Tong University}
  \city{Shanghai}
  \country{China}
  }

\author{Jia Wang}
\affiliation{%
  \institution{Shanghai Jiao Tong University}
  \city{Shanghai}
  \country{China}
}

\author{Zicheng Zhang}
\affiliation{%
  \institution{Shanghai Jiao Tong University}
  \institution{Shanghai AI Laboratory}
  \city{Shanghai}
  \country{China}
}

\author{Guangtao Zhai}
\affiliation{%
  \institution{Shanghai Jiao Tong University}
  \institution{Shanghai AI Laboratory}
  \city{Shanghai}
  \country{China}
}


\renewcommand{\shortauthors}{Xiaorong Zhu et al.}

\begin{abstract}
The rapid evolution of Multi-modality Large Language Models (MLLMs) is driving significant advancements in visual understanding and generation. Nevertheless, a comprehensive assessment of their capabilities, concerning the fine-grained physical principles especially in geometric optics, remains underexplored. To address this gap, we introduce \textbf{GOBench}, the first benchmark to systematically evaluate MLLMs' ability across two tasks: 1) \textbf{Generating Optically Authentic Imagery} and 2) \textbf{Understanding Underlying Optical Phenomena}. We curates high-quality prompts of geometric optical scenarios and use MLLMs to construct GOBench-Gen-1k dataset.
We then organize subjective experiments to assess the generated imagery based on \textit{Optical Authenticity}, \textit{Aesthetic Quality}, and \textit{Instruction Fidelity}, revealing MLLMs' generation flaws that violate optical principles. For the understanding task, we apply crafted evaluation instructions to test optical understanding ability of eleven prominent MLLMs. 
The experimental results demonstrate that current models face significant challenges in both optical generation and understanding. The top-performing generative model, GPT-4o-Image, cannot perfectly complete all generation tasks, and the best-performing MLLM model, Gemini-2.5Pro, attains a mere 37.35\% accuracy in optical understanding. Database and codes
are publicly available at https://github.com/aiben-ch/GOBench.
\end{abstract}



\begin{teaserfigure}
  \centering
\vspace{-4pt}
\includegraphics[width=0.97\linewidth]{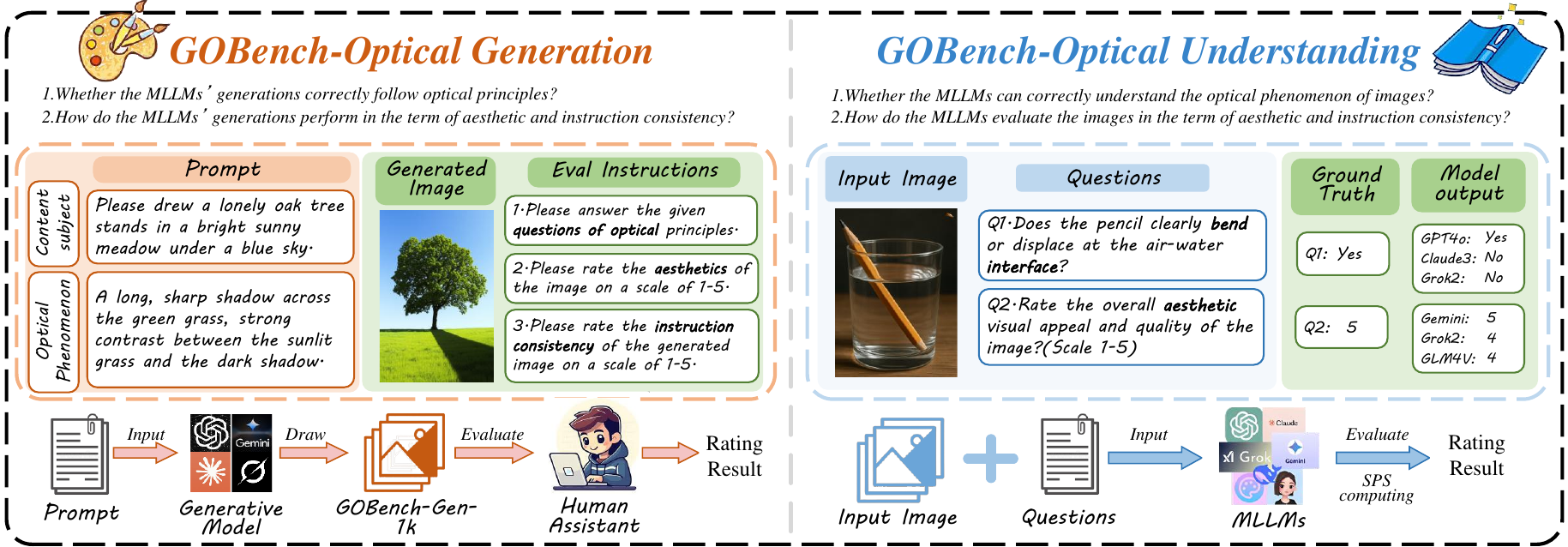}
\vspace{-6pt}
\caption{We propose \textbf{GOBench}, the first benchmark on emerging abilities of MLLMs on geometric optical generation and understanding. On the left side, GOBench-Optical Generation is to judge the MLLM's generations, termed as GOBench-Gen-1k, whether follow optical principles; on the right side, GOBench-Optical Understanding is to judge if MLLMs can correctly understand the optical phenomenon of images.}
\label{overview}


\end{teaserfigure}

\definecolor{mycolor_green}{HTML}{E6F8E0}
\maketitle
\thispagestyle{plain}

\section{Introduction}



\label{sec:introduction}

The advent of large language models (LLMs) such as ChatGPT and Gemini~\citep{floridi2020gpt, team2023gemini}, and their open-source counterparts such as LLaMA~\citep{meta2024introducing}, has driven artificial intelligence (AI) towards general-purpose assistance~\citep{guidetti2019artificial, crafts2021artificial}. Building on the LLM progress, multi-modal large language models (MLLMs)~\citep{bi2024deepseek, liang2024survey} also excel at high-level semantic tasks~\citep{xiao2023multimodal, wang2024llm,hong2025eager,hu2024enhancing}, visual understanding tasks ~\citep{liu2023visual,bai2023qwen,chen2024far,luo2024mono,hutchinson2024llm,hu2024bliva} and multi-modal generation tasks~\citep{podell2023sdxl,rombach2022high,betker2023improving,chen2024multi,mohsin2025retrieval}. Although recent MLLMs master at various high-level multi-modal tasks ~\citep{lu2023chameleon, wang2024emu3, liu2023one},
their proficiency with fine-grained physical principles, for example, the generation and understanding of geometric optics, remains unverified.

Geometric optics, encompassing optical phenomena governed by the Law of Rectilinear Propagation, the Law of Reflection, and Snell's Law, is fundamental to visual appearance and composes most optical scenarios.~\citep{loshin2015geometrical, keating1988geometric}. However, achieving physically plausible optical generation~\citep{tewari2020state} and optical understanding~\citep{zhao2024llm, barron2012shape} is a significant challenge for MLLMs~\citep{fu2024blink,uddin2025ai, li2024eagle}. Current models often produce visually appealing yet physically inaccurate results~\citep{gani2023llm,cherian2024llmphy,abshari2024llm}, with inconsistent lighting or incorrect optical effects~\citep{zhang2025scaling}, limiting their use in high-fidelity applications like realistic content creation or simulation~\citep{hsu2024autovfx,zhao2025envisioning,zhang2025simulation,zheng2025teaching, gu2024your}. This deficiency stems partly from a lack of benchmarks systematically probing these specific optical effects, unlike benchmarks for semantic understanding~\citep{hu2021towards, ma2024groma,aibench} or general VQA~\citep{antol2015vqa}.


To address this gap, we introduce \textbf{GOBench} (\underline{G}eometric \underline{O}ptics Generation and Understanding \underline{Bench}mark), the \textbf{first} benchmark dedicated to systematically evaluating MLLM ability of generating optical authentic imagery and understanding optical phenomena. 
In this benchmark, we constructed the \textbf{GOBench-Gen-1K} image dataset. This involve $340$ unique scenarios focused on key geometric optics phenomena: $108$ for \textit{Direct Light}, $121$ for \textit{Reflection}, and $111$ for \textit{Refraction}. For each scenario, images are then generated by three distinct state-of-the-art MLLMs (GPT-4o-Image, Seedream 3.0, and Imagen 3), and we manually filter out similar or poor quality generations, finally constructed 1k high quality geometric optical images, termed as \textbf{GOBench-Gen-1K}.

For evaluation, GOBench employs a dual-pronged approach focusing on both the fidelity of optical generation task and the depth of optical understanding task. 
Firstly, for \textbf{optical generation assessment}, we conduct a human-labeling experiment where $6$ experts meticulously scored each image across three key dimensions: \textit{Optical Authenticity} (a 0-5 score reflecting physical plausibility based on five specific questions), \textit{Aesthetic Quality} (a 1-5 rating of visual appeal), and \textit{Instruction Fidelity} (a 1-5 rating of adherence to the generation prompt). 
Secondly, for \textbf{optical understanding evaluation}, we benchmark a cohort of 11 prominent MLLMs by tasking them to assess the GOBench-Gen-1K images along these same three dimensions. Their performance is quantified against the human ground truth primarily using a Scaled Proximity Score (SPS), which measures the alignment of their evaluative judgments with human expert consensus.

Experiment results reveal that while current MLLMs' generations attain decent visual quality, they frequently exhibit discernible flaws in adhering to geometric optical principles. And in the optical understanding tasks, even advanced models face substantial challenges in correctly understanding and assessing the optical phenomena in GOBench-Gen-1K. These findings highlight that robust, physically grounded optical generation and understanding remain critical areas for future MLLM development.

In summary, our main contributions are three-fold:
\begin{itemize}
    \item We construct \textbf{Gobench-Gen-1k}, a first-of-its-kind dataset of 1k images specifically designed to test MLLMs on diverse geometric optics scenarios.
    \item We establish a rigorous benchmark for \textbf{optical generation assessment}, providing a gold standard for evaluating fidelity of MLLMs' generations concerning geometric optics principles.
    \item We define a systematic benchmark for the \textbf{optical understanding capabilities} of MLLMs, using a Scaled Proximity Score against human ground truth to quantify alignment with expert judgment. The overview  of our work are summarized in Figure~\ref{overview}.
\end{itemize}

\section{Constructing the GOBench}

\begin{figure}[t]
  \centering
  \includegraphics[width=0.40\textwidth]{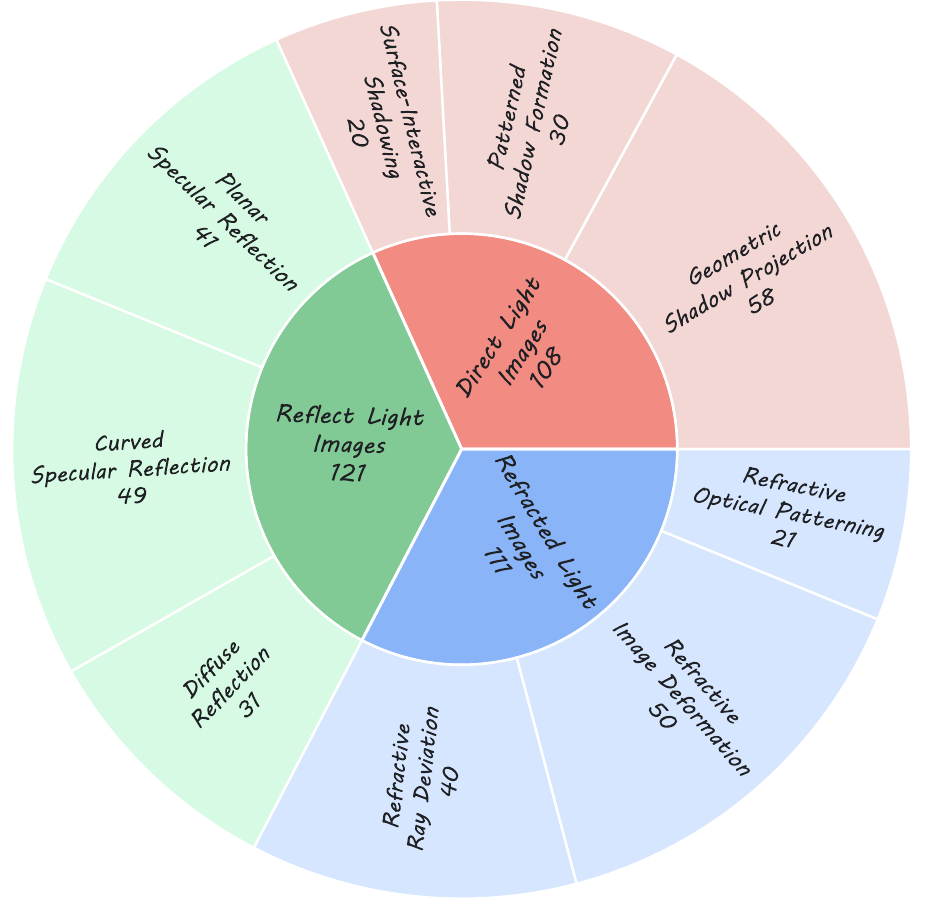} 
  \vspace{-8pt}
  \caption{\textbf{Task Distribution of GOBench-Gen-1K,} involves three main optical categories: \textit{Direct light}, \textit{Reflect light}, and \textit{Refracted light}. Each category includes various subcategories, facilitating a comprehensive dataset.}
  \vspace{-12pt}
  \label{fig:pie}
\end{figure}

\begin{figure*}[t]
  \centering
  \vspace{-2pt}
  \includegraphics[width=0.97\textwidth]{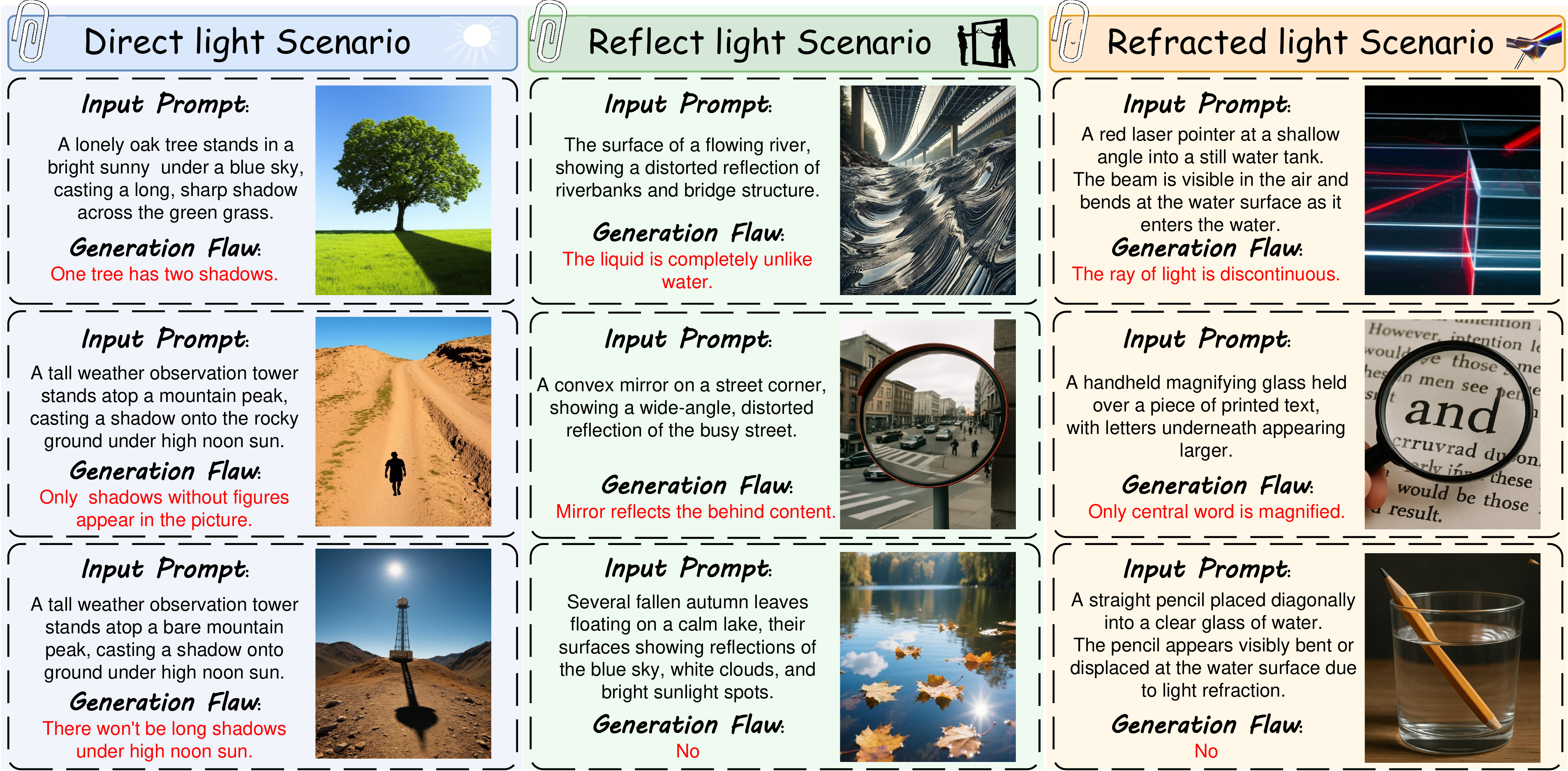} 
  \vspace{-3pt}
  \caption{\textbf{Examples of \textit{GOBench-Gen-1k} that show cases of designed scenarios, including direct light scenario, reflect light scenario and refracted light scenario. Each case includes the input prompt, and the output image generated by state-of-art MLLM. The red words represent the obvious flaws of the MLLM's generations that violating optical or basic physical principles.}}
  \label{fig:dataset}
  \vspace{-12pt}
\end{figure*}

While current Multi-modal Large Language Model (MLLM) benchmarks often assess high-level semantic understanding, the evaluation of MLLMs on fine-grained physical principles, particularly fundamental geometric optics, remains underexplored. Humans readily understand how light interacts with the world—propagating directly, reflecting off surfaces, and refracting through media. However, instilling this nuanced understanding of geometric optics into MLLMs for both accurate image generation and robust understanding presents significant challenges.

To objectively assess current MLLM capabilities in this domain and identify their limitations, we introduce GOBench (Geometric Optics generation and understanding Benchmark). The construction of GOBench involves three core components:

(1) GOBench-Gen-1k Construction: We develope GOBench-Gen-1k, a novel dataset of 1k images. This dataset consists of various geometric optical scenarios in three foundational categories: Direct Light, Reflection, or Refracted Light. Each scenario is rendered by three distinct state-of-the-art MLLMs (GPT-4o-Image, SeeDream 3.0, and Imagen 3) from detailed textual prompts.

(2) Optical Generation Assessment: To establish a human ground truth for generation quality, six experts are invited to score each of the generated images. This subjective evaluation experiment assessed Optical Authenticity, Aesthetic Quality, and Instruction Fidelity, providing a gold standard for how accurately the generative models depicted the intended optical effects.

(3) Optical Understanding Evaluation: To evaluate LMMs' capacity to interpret depicted optical phenomena, we benchmark 11 prominent MLLMs. These models are tasked to evaluate the \textit{GOBench-Gen-1k} across the same three dimensions (Optical Authenticity, Aesthetic Quality, Instruction Fidelity), with their performance measured against the human ground truth.

The subsequent subsections will detail the methodologies employed in the construction of the \textit{GOBench-Gen-1k} and the design of our systematic evaluation framework for both optical generation and understanding.

\subsection{GOBench-Gen-1k Construction}

The construction of the GOBench dataset, termed as GOBench-Gen-1k, involves a two-stage process: comprehensive scenario design followed by systematic image generation. Initially, we curate 340 unique scenarios, which target three major categories of geometric optics phenomena: \textit{Direct Light}, \textit{Reflection}, and \textit{Refraction}. The distribution of the designed scenarios across the three main optical categories is presented in Figure~\ref{fig:pie}. 
Each scenario comprises a detailed textual prompt and a specific set of "Optical Authenticity" questions 
tailored to the phenomenon, intended to guide both optical generation and understanding evaluation. The characteristics of the scenarios within each category is detailed as follows.

\textbf{Direct Light.}
Following the law of rectilinear propagation, direct light scenarios demonstrate light propagation from a source and the consequent formation of shadows. Accurately depicting direct light and shadows demands models to go beyond simple illumination, incorporating an implicit understanding of how light interacts with occluders to create geometrically sound and contextually appropriate visual effects. We define three representative subcategories:
\textit{1) Geometric Shadow Projection}, emphasizing the accurate shape, length, and orientation of shadows based on light source and occluder form.
\textit{2) Patterned Shadow Formation}, examining shadows that create intricate texture or patterns due to the occluder's structure.
\textit{3) Surface-Interactive Shadowing}, investigating how shadow appearance is affected by the receiving surface's physical properties or atmospheric conditions.
Together, these scenarios provide a comprehensive testbed for evaluating a model’s capability to intelligent and render direct illumination, shadow casting and the nuanced visual interplay between light, occluders, and diverse environmental surfaces.

\textbf{Reflect Light.}
Obeying the law of reflection, reflect light cases evaluate a model’s proficiency in generation how light bounces off various surfaces, leading to visual effects ranging from clear mirror-like images to diffuse sheens. Understanding and accurately depicting reflection is essential for conveying material properties, surface characteristics, and the richness of environmental context within generated imagery. The scenarios are constructed to probe a model's grasp of reflection by encompassing three key aspects:
\textit{1) Planar Specular Reflection}, which focuses on the generation of clear, largely undistorted mirror images from flat, smooth surfaces such as calm water or clean glass, emphasizing high-fidelity mirroring.
\textit{2) Curved Specular Reflection}, which examines the generation of predictable image distortions, magnification or minification effects.
\textit{3) Diffuse Reflection}, addressing the scattering of light from surfaces that are inherently rough, textured, or wet, resulting in a loss of sharp specular highlights and the appearance of generalized blurred reflections rather than distinct images.
This category require models to exhibit implicit knowledge of accurate specular mirroring, complex distortions from curved surfaces, and the subtle interplay of light with varied material textures and conditions.

\textbf{Refracted Light.}
Complying with the Snell's Law, refracted light cases assess a model’s proficiency in generation the deflection of light as it traverses interfaces between different transparent media. This optical phenomenon is fundamental to a multitude of visual effects, and its accurate depiction by MLLMs requires an implicit understanding of how variations in refractive indices modify light paths, leading to observable outcomes such as image displacement, deformation, or chromatic dispersion. Our scenario construction for this category systematically probes these capabilities via three principal aspects:
\textit{1) Refractive Ray Deviation}, concentrating on the precise generation of light ray paths. These paths must exhibit accurate angular changes at media interfaces, in accordance with optical laws.
\textit{2) Refractive Image Deformation}, which concerns the faithful representation of alterations to the perceived visual form of objects or backgrounds. Such alterations include bending, displacement, or magnification when viewed through transparent refractive elements or due to atmospheric optical effects.
\textit{3) Refractive Optical Patterning}, which explores the model's ability to simulate phenomena where refraction organizes light into structured visual patterns or chromatic displays, encompassing effects like caustics, spectral dispersion, and birefringence.

These subcategories, spanning diverse refractive effects, provide a detailed assessment of an MLLM's capacity for simulating complex light-media interactions and its understanding of resultant visual phenomena.

Following the above designed scenarios, we generate images with three state-of-the-art Large multi-modality models: GPT-4o-Image, SeeDream3.0, and Imagen 3. Each model generates one image per scenario; we filter out some cases that exhibit extremely poor generation quality and those that show excessive content similarity, ultimately constructing GOBench-Gen-1k with total 1k cases. And examples of GOBench-Gen-1k are demonstrated in Figure~\ref{fig:dataset}. It can be seen that the MLLMs' generations have decent quality but there are still obvious flaws that violate the optical or basic physical principles.

\subsection{Benchmark on Optical Generation}

To establish a robust ground truth for the optical generation quality of the MLLMs, we conduct a human expert evaluation experiment. A panel of six human experts is invited to meticulously score the designed 1k cases generated by by GPT-4o-Image, SeeDream 3.0, and Imagen 3.
The evaluation is performed across three key dimensions for each case:
\textit{1). Optical Authenticity}:This dimension critically assesses the physical plausibility and accurate depiction of the target geometric optical phenomenon (Direct Light, Reflection, or Refraction) within the generated image. 
For each of the scenarios, experts are guided by a detailed, phenomenon-specific rubric consisting of five "Optical Authenticity" questions. 
These questions are carefully designed to cover both general principles applicable to each main optical category and nuances specific to individual scenarios.

\begin{figure}[t]
  \centering
  
  \includegraphics[width=0.48\textwidth]{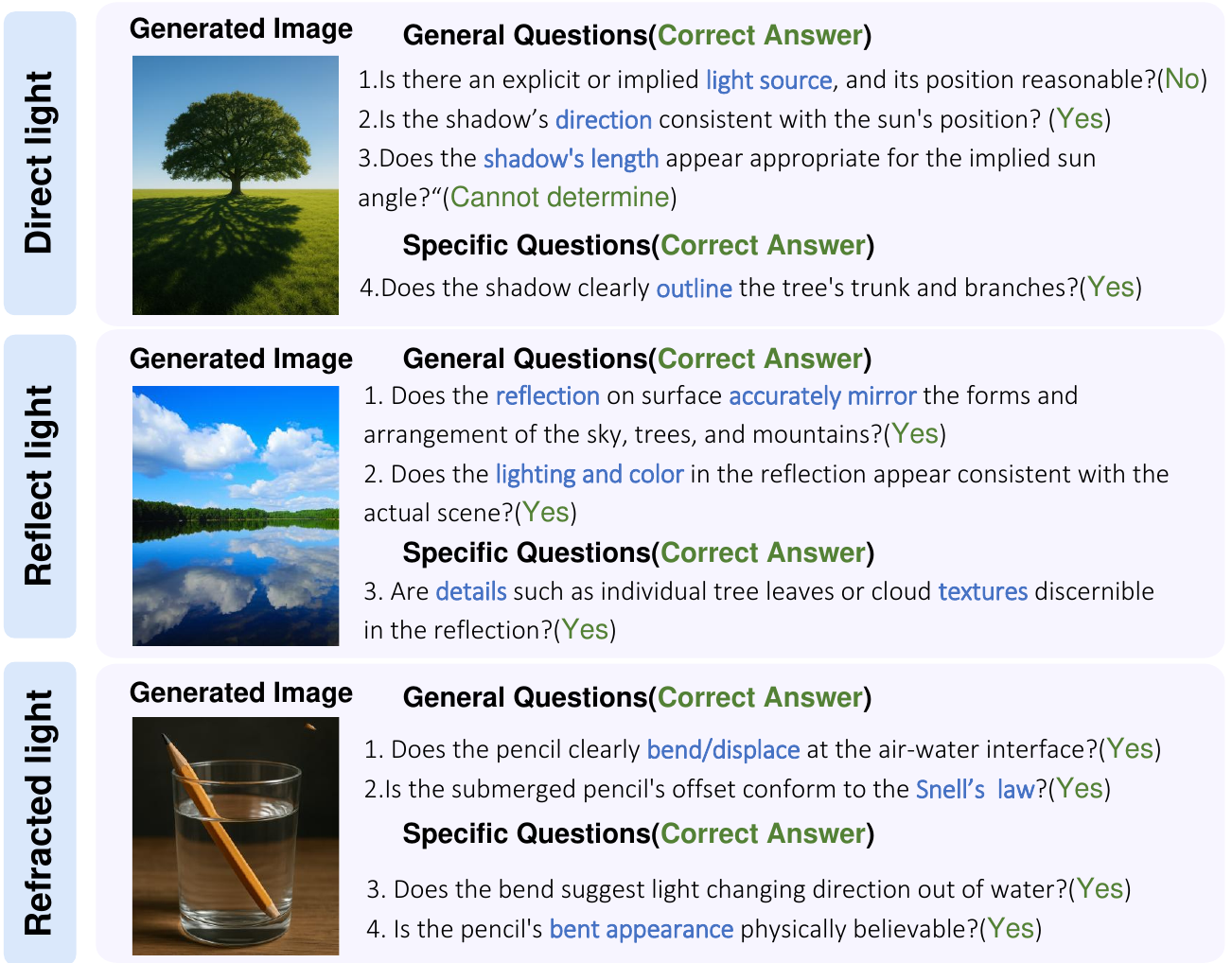} 
  \vspace{-12pt}
  \caption{\textbf{Optical Authenticity questions of GOBench.} The questions based on \textit{GOBench-Gen-1k} that evaluates the Optical Authenticity of MLLM's generations, and each case contains general and specific questions. }
  \vspace{-8pt}
  \label{fig:authenticity_questions}
  \vspace{-2pt}
\end{figure}

Generally, two to three questions for each case probe category-specific principles. For instance, Direct Light scenarios consider questions about the plausible existence of light sources, the consistency of shadow direction with the light source, and the appropriateness of shadow length relative to the implied light source angle. Reflection scenarios assess if reflections accurately mirror the form and arrangement of original objects and if lighting, color and shape are consistent and reasonable between the scene and its reflection. Refraction scenarios typically query the obvious bending or displacement of objects and light paths at media interfaces and the plausibility of the refractive angles.
The remaining two to three questions for each case delve into scenario-specific nuances to further test the authenticity of the optical generation. These questions are designed to assess the illumination characteristics, shadow outline, and contextual details as they manifest under the specific optical phenomenon.
Illustrative examples of these category-specific and scenario-specific questions are provided in Figure~\ref{fig:authenticity_questions}.

\begin{figure*}[htbp]
  \centering
  \includegraphics[width=0.98\textwidth]
  {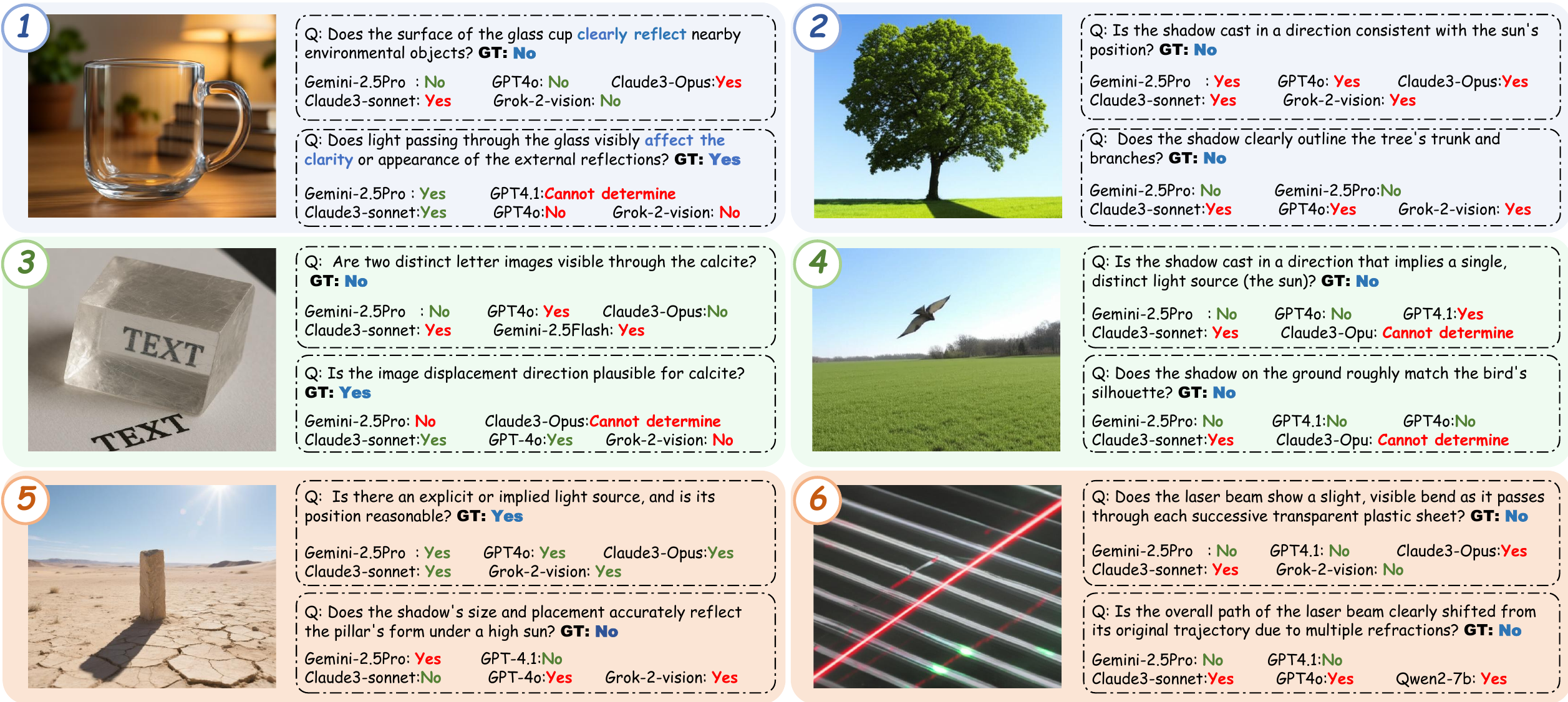} 
  \vspace{-2pt}
  \caption{\textbf{Examples of several different models’ answers to the optical authentic questions.}GT represents ground truth result; the Green words represent correct answers and the red words represent the wrong answers. }
  \vspace{-8pt}
  \label{fig:MLLMassess}
\end{figure*}

Each of these five guiding questions for a given scenario is answerable with "Yes" (contributing 1 point), "No" (0 points), or "Cannot be determined" (0.5 points). The "Yes" or "No" responses reflect a judgment on whether the depicted optical phenomenon is reasonably portrayed based on the available visual information within the image. For instance, a "No" answer should be given to the first question for the direct light case demonstrated in Figure~\ref{fig:authenticity_questions}, which queries light source plausibility. The reason is that the image shows a tree shadow cast to the left while the sky exhibits uniform brightness without any visual cue to suggest a correspondingly positioned light source. The correct scenario should demonstrate a brighter region on the right.
The "Cannot be determined" option (0.5 points) is specifically assigned when the optical phenomenon itself appears plausible, but the image lacks sufficient contextual information to definitively affirm or negate the specific query. For instance, this would apply if a question assesses shadow direction consistency, but the image depicts the shadow without clearly showing the light source or the occluding object, resulting in the difficulty to make a conclusive judgment on the question. Finally, the sum of these points results in an Optical Authenticity score ranging from 0 to 5 for each image.
\textit{2). Aesthetic Quality}: Experts rate the overall visual appeal, composition, and artistic merit of each image on a 1-5 scale (1 being lowest, 5 being highest), based on the question: "Please rate the aesthetics of the image on a scale of 1-5."
\textit{3). Instruction Fidelity}: The adherence of the generated image to the original textual prompt, particularly in its success at showcasing the intended optical phenomenon, is also rated on a 1-5 Likert scale (1 being lowest, 5 being highest), guided by the question: "Based on the image and the provided prompt, please rate the instruction fidelity of the generated image on a scale of 1-5."

Finally, the averaged expert scores for these three dimensions are applied to evaluate the MLLMs' ability of optical generation, which also serve as the definitive ground truth for the following understanding tasks.

\vspace{-8pt}
\subsection{Benchmark on Optical Understanding}

In the second primary task of GOBench, we assess the capacity of multi-modality large language models (MLLMs) to understand the underlying optical phenomena depicted in the \textit{GOBench-Gen-1k}.  Our methodology for this assessment involves two key stages: 1) tasking the MLLMs to provide evaluative judgments across Optical Authenticity, Aesthetic Quality, and Instruction Fidelity for each image based on structured prompts, and 2) subsequently defining and applying clear quantitative metrics to measure the performance of these MLLM evaluations against human ground truth.

The evaluative framework leverages the 1k image instances from \textit{GOBench-Gen-1k}. For each image, MLLMs are provided with the image, its original English textual generation prompt, and five scenario-specific "Optical Authenticity" questions.
A system prompt directs MLLMs to act as expert image critics, performing three sub-tasks: (1)~answering each Optical Authenticity question ("Yes", "No", or "Cannot be determined"), from which an authenticity score (0-5 scale) is computed; (2)~assigning an \textit{Aesthetic Quality} rating (1-5 scale, Poor to Excellent); and (3)~providing an \textit{Instruction Fidelity} rating (1-5 scale) reflecting alignment with the generation prompt, especially concerning intended optical effects. 


We assesse 11 MLLMs via automated Python scripts with standardized parameters.
(e.g, temperature 0.1). Specifically, the different answers of MLLMs to the Authenticity is shown in Figure~\ref{fig:MLLMassess}. The MLLMs can answer correctly in simple questions such as the recognition of light source as shown in the fifth case. However, for interfering questions, only outstanding models like Gemini-2.5-Pro and GPT-4o, can provide correct answer as shown in the fourth case.

We also compare the MLLM's score ($S_{MLLM}$) against the corresponding human ground truth score ($S_{GT}$). We define a maximum permissible absolute difference, $\delta_{max}$ (set to 0.5), for assigning partial credit. Scores are linearly scaled within this difference, yielding the \textbf{Scaled Proximity Score} ($SPS$) for each case:

\begin{equation}
SPS_{\text{case}} = 
\begin{cases} 
1 - \frac{|S_{MLLM} - S_{GT}|}{\delta_{max}} & \text{if } |S_{MLLM} - S_{GT}| \leq \delta_{max} \\
0 & \text{if } |S_{MLLM} - S_{GT}| > \delta_{max}
\end{cases}
\label{eq:scaled_proximity_score}
\end{equation}

The final reported $SPS$ for each dimension (Optical Authenticity, Aesthetic Quality, and Instruction Fidelity) is the average of these $SPS_{\text{case}}$ values. This metric offers a fine-grained performance measurement of MLLM alignment with ground truth.

\section{Experiments}
\label{sec:experiments}

To comprehensively evaluate MLLM capabilities in the domain of geometric optics, GOBench facilitates several distinct experimental analyses. First, we examine the characteristics of our human expert evaluation framework itself by assessing the inter-dimensional correlation of the defined rating criteria. Second, we quantify the optical generation proficiency of the state-of-the-art generative models. Finally, our investigation benchmarks the optical understanding and evaluative performance of a diverse cohort of 11 MLLMs. The subsequent subsections detail the setup and present the findings from each of these analytical components.

\subsection{Validity of Evaluation Dimension}
\label{sec:validity_eval_dimension}


\begin{table}[t]\small
    \centering
    \renewcommand\arraystretch{1}
    \renewcommand\tabcolsep{10pt} 
    \caption{Pairwise Correlation Coefficients (PLCC and SRCC) between Rating Dimensions}
    \vspace{-5pt}
    \begin{tabular}{l|cc} 
    \hline
    \multicolumn{1}{l|}{\textbf{Comparison Pair}} & \textbf{PLCC} & \textbf{SRCC} \\ \hline
        Aesthetics vs Authenticity  & 0.2213 & 0.2085 \\
        Authenticity vs Fidelity  & 0.2942 & 0.2716 \\
        Aesthetics vs Fidelity   & 0.7036 & 0.6982 \\
       \hline
    \end{tabular}
    \label{tab:1.dime_corre} 
\end{table}

To validate the rationality of our designed evaluation dimensions, we first analyze the pairwise correlation between the human expert ratings for Optical Authenticity, Aesthetic Quality, and Instruction Fidelity across the GOBench-Gen-1k cases. Table~\ref{tab:1.dime_corre} presents the Pearson Linear Correlation Coefficient (PLCC) and Spearman Rank Correlation Coefficient (SRCC) for these comparisons.

The results reveal a low correlation between Aesthetic Quality and Optical Authenticity (PLCC = 0.2213, SRCC = 0.2085), and similarly between Instruction Fidelity and Optical Authenticity (PLCC = 0.2942, SRCC = 0.2716). This indicates that human evaluators assess the physical correctness of optical phenomena largely independently of an image's general visual appeal. And the correlation between Aesthetic Quality and Instruction Fidelity (PLCC = 0.7036, SRCC = 0.6982) still shows that the results in each dimension are largely independent. The generally low inter-correlations reveal that each dimension effectively captures distinct aspects of image quality and correctness concerning geometric optics.

\subsection{Results on \textbf{Optical Generation}}
\label{sec:results_generation}



\begin{table}[t]\small 
    \centering
    \renewcommand\arraystretch{1}
    \renewcommand\tabcolsep{6pt} 
    \vspace{-3pt}
    \caption{Average Evaluation Scores of MLLMs' generations}
    \vspace{-5pt}
    \begin{tabular}{l|ccc} 
    \hline
    \multicolumn{1}{l|}{\textbf{LMMs}} & \textbf{Aesthetics} & \textbf{Fidelity} & \textbf{Authenticity} \\ \hline
        GPT-4o-Image      & \underline{3.95} & \textbf{3.97} & \textbf{4.09} \\
        Seedream3.0 & \textbf{4.02}   & \underline{3.94} & \underline{4.02} \\
        Imagen 3   & 3.70            & 3.67          & 3.48          \\
       \hline
    \end{tabular}
    \vspace{-5pt} 
    \label{tab:2.result_render} 
\end{table}



To evaluate the optical generation proficiency of the three generative MLLMs, we compute average human expert scores across three dimensions, as presented in Table~\ref{tab:2.result_render}.
Among the generative models, GPT-4o-Image achieved the highest average score for Optical Authenticity (referred to as Authenticity in the table) with 4.09 out of 5, suggesting a relatively strong capability in depicting the core geometric optics phenomena as intended by the scenarios. Seedream 3.0 excelled in Aesthetic Quality, scoring an average of 4.02, and also demonstrated high Instruction Fidelity at 3.94, nearly matching GPT-4o's 3.97 in this regard. Imagen 3 has the smoothest generation, while generally scored lower, particularly in Optical Authenticity (3.48).
While these state-of-the-art models demonstrate the ability to generate images that are often optically plausible and aesthetically acceptable, there are still gaps to achieve highly optical fidelity and instruction consistency. These findings underscore the limitations in current generative models concerning the nuanced and physically accurate generation of geometric optics.

\subsection{Results on \textbf{Optical Understanding}}


\begin{table}[t]\small
    \centering
    \renewcommand\arraystretch{1}
    \renewcommand\tabcolsep{8pt} 
    \caption{\textbf{MLLMs' Optical Understanding Performance.} Scaled Proximity Scores (SPS) of 11 MLLMs against human ground truth for Optical Authenticity, Aesthetic Quality, and Instruction Fidelity on \textit{GOBench-Gen-1k}. Best score of each dimension is emphasized with boldface, second best is underlined.}
    \vspace{-5pt} 
    \resizebox{\linewidth}{!}{\begin{tabular}{l|ccc}
    \hline
    \multicolumn{1}{l|}{\textbf{Categories}} & \multicolumn{3}{c}{\textbf{GOBench-Understanding}} \\
    \cdashline{1-1} \cdashline{2-4}
    \multicolumn{1}{l|}{\textbf{MLLMs}} & \textit{Authenticity$\uparrow$} & \textit{Aesthetics$\uparrow$} & \textit{Fidelity$\uparrow$} \\ \hline 

        Gemini-2.5Pro (\textit{0506})           & \textbf{37.35\%}    & 18.78\%             & 8.31\%  \\ 
        Claude3-Opus                          & \underline{33.93\%} & \underline{31.23\%} & \textbf{28.29\%} \\ 
        GPT4o (\textit{latest})                 & 32.63\%             & 8.17\%              & 6.04\%  \\
        GPT4\_1 (\textit{20250414})              & 32.23\%             & 5.74\%              & 5.37\%  \\
        GPT4\_1\_mini (\textit{20250414})         & 31.93\%             & 9.24\%              & 6.64\%  \\
        Claude3-Sonnet                        & 31.53\%             & 15.28\%             & 7.97\%  \\
        Grok-2-Vision~\cite{lee2025bridging}                         & 28.53\%             & 15.48\%             & 7.61\%  \\
        Gemini-2.5Flash (\textit{0417})         & 28.43\%             & 10.84\%             & 6.74\%  \\
        GPT4\_1\_nano (\textit{20250414})         & 28.33\%             & 25.76\%             & 8.48\%  \\
        \cdashline{1-4}
        Glm4v(\textit{9b})~\cite{hong2024cogvlm2}                      & 27.73\%             & \textbf{32.03\%}    & \underline{27.36\%} \\ 
        Qwen2\_5vl(\textit{7b})~\cite{wang2024qwen2}                  & 27.43\%             & 30.16\%             & 24.36\% \\
       \hline
    \end{tabular}}
    \vspace{-12pt}
    \label{tab:3.reason_accu} 
\end{table}


The GOBench also evaluates the optical understanding and evaluative capabilities of 11 MLLMs, by comparing their assessment results of the GOBench-Gen-1k against human ground truth. 
Table~\ref{tab:3.reason_accu} presents the performance using our primary metric, the Scaled Proximity Score (SPS) (Equation~\ref{eq:scaled_proximity_score}), with models sorted by their SPS in the Optical Authenticity dimension.
The results indicate that even advanced MLLMs find optical understanding challenging. Gemini-2.5Pro (0506) leads in Optical Authenticity SPS with 37.35\%, followed by Claude3-Opus (33.93\%) and GPT-4o (latest) (32.63\%). These top scores, considerably below 50\%, highlight the substantial difficulty MLLMs in correctly judge the optical phenomena in the given images. Tops scores for aesthetic quality(32.03\%) and instruction consistency(28.29\%) further reveal the limitation of MLLMs' ability to understand the geometric optical phenomena.
\section{Conclusion}

In this work, we introduce GOBench, a novel benchmark designed to rigorously evaluate Multi-modal Large language Models'(MLLMs) abilities in geometric optical generation and understanding. 
GOBench systematically assesses performance across three core optical categories:Direct Light, Reflection, and Refraction. 
We apply a multi-dimensional framework to assess the models' ability of optical generation and understanding including Optical Authenticity, Aesthetic Quality, and Instruction Fidelity. 
The experiment results reveal that state-of-the-art models still face significant challenges in achieving consistent, physically accurate optical generation and robust optical understanding ability. 
These findings underscore critical areas for future research in enhancing the physical world understanding of MLLMs.


\begin{acks}
This work was supported by the National Natural Science Foundation of China under Grants No. 62225112 and U24A20220.
\end{acks}

\bibliographystyle{ACM-Reference-Format}
\bibliography{main}



\end{document}